# A machine learning model for skillful climate system prediction

Chenguang Zhou[1,2#], Lei Chen[3#], Xiaohui Zhong[3#], Bo Lu[1,2*], Hao Li[3*], Libo Wu[4,5,6*], Jie Wu[1], Jiahui Hu[2,7], Zesheng Dou[2], Pang-Chi Hsu[8,9], Xiaoye Zhang[10]

[1]State Key Laboratory of Climate System Prediction and Risk Management, National Climate Centre, China Meteorological Administration, Beijing, China

[2]Xiong'an Institute of Meteorological Artificial Intelligence, Xiong'an New Area, China

[3]Artificial Intelligence Innovation and Incubation Institute, Fudan University, Shanghai, China

[4]School of Data Science, Fudan University, Shanghai, China.

[5]Institute for Big Data, Fudan University, Shanghai, China.

[6]MOE Laboratory for National Development and Intelligent Governance, Fudan University, Shanghai, China

[7]Xinjiang Climate Center, Urumqi, China

[8]State Key Laboratory of Climate System Prediction and Risk Management/Key Laboratory of Meteorological Disaster, Ministry of Education/Collaborative Innovation Center on Forecast and Evaluation of Meteorological Disasters, Nanjing University of Information Science and Technology, Nanjing, China

[9]School of Atmospheric Sciences, Nanjing University of Information Science and Technology, Nanjing, China

[10]State Key Laboratory of Severe Weather Meteorological Science and Technology, Chinese Academy of Meteorological Sciences, Beijing, China

\# These authors contributed equally to this work.

\* Corresponding author(s). E-mail(s): bolu@cma.gov.cn; lihao_lh@fudan.edu.cn; wulibo@fudan.edu.cn

**Abstract**

Climate system models (CSMs), through integrating cross-sphere interactions among the atmosphere, ocean, land, and cryosphere, have emerged as pivotal tools for deciphering climate dynamics and improving forecasting capabilities. Recent breakthroughs in artificial intelligence (AI)-driven meteorological modeling have demonstrated remarkable success in single-sphere systems and partially spheres coupled systems. However, the development of a fully coupled AI-based climate system model encompassing atmosphere-ocean-land-sea ice interactions has remained an unresolved challenge. This paper introduces FengShun-CSM, an AI-based CSM model that provides 60-day global daily forecasts for 29 critical variables across atmospheric, oceanic, terrestrial, and cryospheric domains. The model significantly outperforms the European Centre for Medium-Range Weather Forecasts (ECMWF) subseasonal-to-seasonal (S2S) model in predicting most variables, particularly precipitation, land surface, and oceanic components. This enhanced capability is primarily attributed to its improved representation of intra-seasonal variability modes, most notably the Madden-Julian Oscillation (MJO). Remarkably, FengShun-CSM exhibits substantial potential in predicting subseasonal extreme events. Such breakthroughs will advance its applications in meteorological disaster mitigation, marine ecosystem conservation, and agricultural productivity enhancement. Furthermore, it validates the feasibility of developing AI-powered CSMs through machine learning technologies, establishing a transformative paradigm for next-generation Earth system modeling.

**Introduction**

Weather forecasting within a two-week horizon primarily depends on atmospheric initial conditions, whereas climate predictions at sub-seasonal and longer timescales require a comprehensive consideration of the climate system's boundary conditions. These boundary forcings encompass three categories: climate variability modes (such as Madden Julian Oscillation (MJO), North Atlantic Oscillation (NAO), and El Niño-Southern Oscillation (ENSO)), slow-varying components in the Earth system (such as upper ocean heat content, sea ice, and soil moisture), and external forcing factors (such as solar radiation, greenhouse gases, and volcanic aerosols). These conditions constitute fundamental sources of predictability for climate forecasts across temporal scales (Rasmusson and Carpenter, 1982; Hurrell and Deser, 2010; Koster et al., 2011; Zuo et al., 2016; Vitart, 2017; Frederiksen et al. 2018; Patterson et al. 2022). Therefore, developing coupled climate system models (CSMs) that integrate multi-sphere interactions is essential for achieving reliable climate predictions. Over decades of development, physics-based CSMs have emerged as powerful tools for climate prediction and effective methods for quantifying cross-sphere interactions and their climatic impacts (Vitart and Robertson, 2018; Pegion et al., 2019; Wang et al. 2009; Ren et al., 2019; Wu et al., 2024). Nevertheless, the conventional physics-driven paradigm remains constrained by prohibitive computational costs and temporal inefficiency, which hinders progress in enhancing spatial resolution and integrating sophisticated physical parameterizations.

In recent years, artificial intelligence (AI)-driven large models have demonstrated groundbreaking advancements in weather forecasting, including some notable examples such as Pangu-Weather, FuXi, GraphCast, NowcastNet, AIFS, NeuralGCM, and GenCast (Bi et al., 2023; Chen et al., 2023; Lam et al., 2023; Zhang et al., 2023; Kochkov et al., 2024; Lang et al., 2024; Price et al., 2024). Compared to conventional physics-based models, AI-based large models exhibit superior predictive skills, unparalleled computational efficiency, and significantly reduced operational costs, thereby establishing novel methodologies for reliable weather forecasting. However, these weather-oriented AI models remain inadequate for subseasonal-to-seasonal (S2S) and longer-term climate predictions (He et al., 2021; Weyn et al., 2021). As the first AI-based subseasonal prediction model, FuXi-S2S achieves atmosphere-ocean coupling through machine learning techniques. The model outperforms the European Centre for Medium-Range Weather Forecasts (ECMWF) S2S model in subseasonal predictions, particularly in forecasting the MJO and precipitation anomalies. However, the absence of critical predictability sources—including upper-ocean heat content, land surface processes, and sea ice dynamics—constrains FuXi-S2S's ability to further improve its prediction accuracy and achieve reliable longer-term climate predictions. Despite recent studies have reported several AI-based climate prediction models, they either incorporate limited spheres of the Earth system or focus on specific regions (Cresswell-Clay et al., 2024; Ling et al., 2024; Wang et al., 2024; Watt-Meyer et al., 2024; Kent et al., 2025; Mu et al., 2025), rather than representing conventional comprehensive multi-sphere coupled CSMs. To date, an AI-based multi-sphere coupled CSM encompassing atmosphere-ocean-land-sea ice interactions has yet to be developed.

To overcome these challenges, we present the FengShun-CSM model, a novel AI-driven CSM that constitutes a breakthrough in multi-sphere coupling. FengShun-CSM achieves full atmosphere-ocean-land-sea ice coupling through machine learning techniques, capable of producing 60-day global forecasts with a daily temporal resolution and a spatial resolution of 0.25°. This model integrates 29 variables, including 17 atmospheric variables, 4 land variables, 6 oceanic variables, and 2 sea ice variables. Additionally, it features an innovative coupling module specifically designed to facilitate the exchange of feature information between the atmospheric and oceanic ensemble models. The architecture of the model is illustrated in Figure 6, and more details about the FengShun-CSM model are available in "Methods".

For a long time, the ECMWF S2S model has been regarded as one of the most outstanding S2S forecasting models (de Andrade et al., 2019; Domeisen et al., 2022). Our results demonstrate that FengShun-CSM outperforms the ECMWF S2S model in predicting multiple variables related to the atmosphere, ocean, land, and sea ice, whether in deterministic, probabilistic, or extreme forecasting. This superior performance is related to FengShun-CSM's advanced skills in predicting intra-seasonal variability modes. Compared to the ECMWF S2S model, FengShun-CSM significantly enhances the prediction skills of the MJO and NAO, thereby also improving its sub-seasonal prediction ability for climate anomalies in mid-to-high latitude regions. For example, FengShun-CSM successfully predicted extreme events such as the cold wave and heatwave in North America and the floods in Europe that occurred in 2021. In addition to its superior accuracy, the computational efficiency of FengShun-CSM also demonstrates significant potential for disaster prevention and mitigation efforts.

**Results**

This study utilizes all testing data from 2021 to systematically evaluate the deterministic, probabilistic, and extreme prediction skills of the FengShun-CSM model for key climatic elements across the atmosphere, ocean, land, and sea ice. The evaluation also includes its ability to predict major intra-seasonal variability modes such as the MJO and NAO, as well as its performance in forecasting extreme events like cold waves and heatwaves. Additionally, the study analyzes the coupling relationships between different spheres in the FengShun-CSM model, with a particular focus on the coupling performance of atmosphere-ocean, atmosphere-land, and atmosphere-ice interactions. To enhance the credibility of FengShun-CSM's prediction skills, its results are compared with those of ECMWF S2S forecasts from the model cycle C47r2 over the same period. Additional evaluations, such as climatological comparisons and an analysis of the European floods caused by an extreme precipitation event in July 2021 (Tuel et al., 2022), are available in the Supplementary Material. Before the evaluation of the prediction skill, FengShun-CSM's forecasting performance for climatology is investigated by comparing the results with observations and ECMWF S2S forecasts. As shown in Supplementary Figures 1–3, both FengShun-CSM and the ECMWF S2S model are able to effectively capture the spatial distribution characteristics of the observed climatology, which validates the representation capability of AI-based CSMs for complex climate systems.

*Prediction of climatic elements*

This subsection compares the deterministic, probabilistic, and extreme prediction skills of FengShun-CSM and ECMWF S2S model for major climatic elements across four spheres in the Earth system. Figure 1 shows the globally-averaged and latitude-weighted temporal anomaly correlation coefficients (TCC) of weekly mean anomalies for FengShun-CSM and ECMWF S2S at different lead weeks. The analysis includes 12 variables: total precipitation (TP), 2 m temperature (T2M), geopotential at 500 hPa (Z500), and outgoing longwave radiation (OLR) from the atmosphere; mean ocean potential temperature and salinity above 300 meters (T300 and S300), mixed layer thickness (MLT), and sea surface height (SSH) from the ocean; soil moisture and temperature above 100 cm (SM100 and ST100) from the land; and sea ice concentration (SIC) and thickness (SIT) from the cryosphere. The analysis is based on the average TCC calculated using all test data from 2021. The results show that the ensemble mean forecasts of FengShun-CSM outperform ECMWF S2S for most variables. Specifically, FengShun-CSM forecasts generally demonstrate higher TCC values than ECMWF S2S forecasts for TP and OLR in the atmosphere, as well as for all oceanic and land variables at all lead weeks. However, for T2M and Z500 in the atmosphere and the two sea ice variables (SIC and SIT), the prediction skills of FengShun-CSM are comparable to those of ECMWF S2S, and in some cases, perform worse at multiple lead weeks. Supplementary Figure 4 presents a comparison of the globally-averaged and latitude-weighted root mean square error (RMSE) of the ensemble mean between FengShun-CSM forecasts and ECMWF S2S forecasts for the 12 variables. Consistent with the findings in Figure 1, FengShun-CSM demonstrates superior forecast performance for TP and OLR in the atmosphere, as well as for all oceanic and land elements at all lead weeks, with lower RMSE values than ECMWF S2S. However, for T2M, Z500, SIC, and SIT, FengShun-CSM shows RMSE values similar to those of ECMWF S2S.

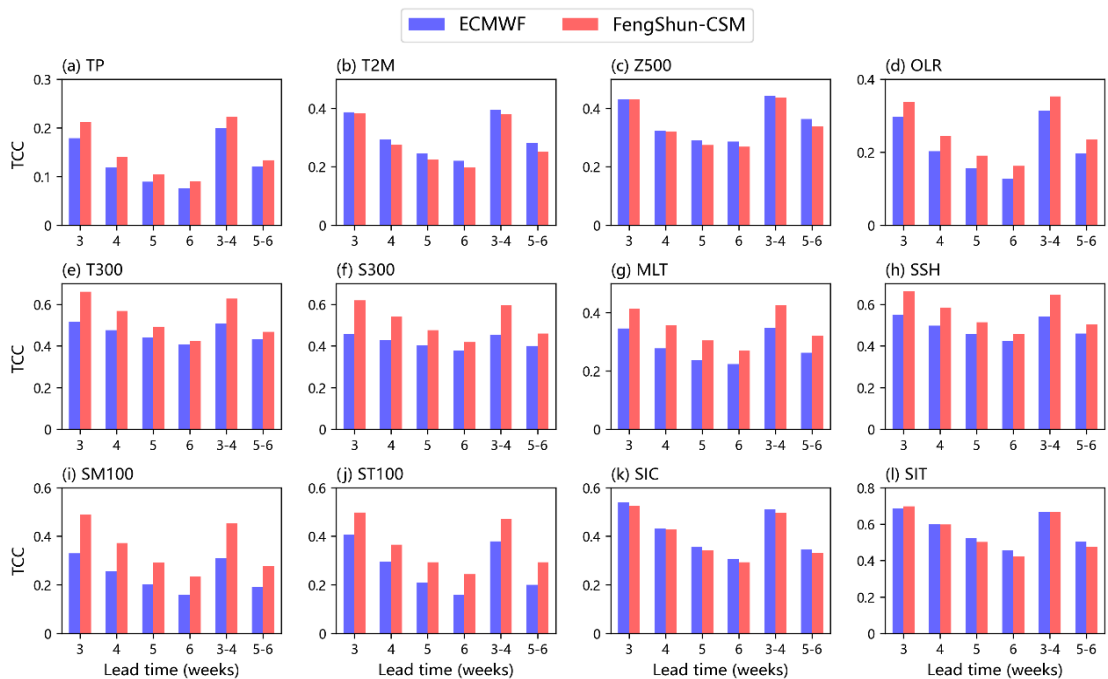

**Figure 1:** Comparison of globally-averaged and latitude-weighted temporal anomaly

correlation coefficient (TCC) of the ensemble mean between ECMWF subseasonal-to-seasonal (S2S) forecasts (in blue) and FengShun-CSM forecasts (in red) for total precipitation (TP), 2 m temperature (T2M), geopotential at 500 hPa (Z500), outgoing longwave radiation (OLR), mean ocean potential temperature above 300 m (T300), mean salinity above 300 m (S300), mixed layer thickness (MLT), sea surface height (SSH), soil moisture from 0 to 100 cm (SM100), soil temperature from 0 to 100 cm (ST100), sea ice concentration (SIC), and sea ice thickness (SIT), using all testing data from 2021.

Supplementary Figures 5 and 6 respectively display the comparison of the globally-averaged and latitude-weighted probabilistic prediction skills, evaluated using the ranked probability skill score (RPSS) metric, and extreme prediction skills for exceeding the 90th climatological percentile, assessed using the Brier Skill Score (BSS) metric, between FengShun-CSM forecasts and ECMWF S2S forecasts. In line with the conclusions on deterministic prediction skills in Figure 1, FengShun-CSM exhibits higher probabilistic and extreme prediction skills than ECMWF S2S for variables with higher deterministic forecasting performance, particularly for oceanic and land variables. Notably, FengShun-CSM demonstrates better extreme prediction skills for most variables, including Z500 and SIT, compared to ECMWF S2S (as shown in Supplementary Figure 6), indicating its potential for more reliable forecasting of extreme events at the sub-seasonal timescale.

Figure 2 shows the spatial distributions of the differences in TCC, RPSS, and BSS between FengShun-CSM forecasts and ECMWF S2S forecasts for anomalies of TP, T300, SM100, and SIC at lead times of 3–6 weeks. It is important to note that this subsection only presents the prediction skills for one variable from each sphere of the Earth system, while the prediction skills for the other eight variables can be found in the supplementary materials (Supplementary Figures 7 and 8). In Figure 2 and Supplementary Figures 7 and 8, red (blue) grid points indicate that FengShun-CSM forecasts performs better (worse) than ECMWF S2S forecasts. For the prediction of multiple variables, such as TP, OLR, T300, S300, MLT, SM100, and ST100, FengShun-CSM demonstrates nearly global positive skills differences, suggesting that it is more proficient in predicting these variables compared to ECMWF. However, for the prediction of other variables, the performance differences between the two models are influenced by regional factors. Interestingly, FengShun-CSM exhibits significantly higher TCC, RPSS, and BSS skills for T2M predictions over most land areas, particularly in monsoon-influenced regions such as Asia, North America, and southern Africa, which are regions where humans reside and where extreme events frequently occur. This highlights the potential of FengShun-CSM in accurately predicting extreme high and low temperature events across most monsoon regions globally.

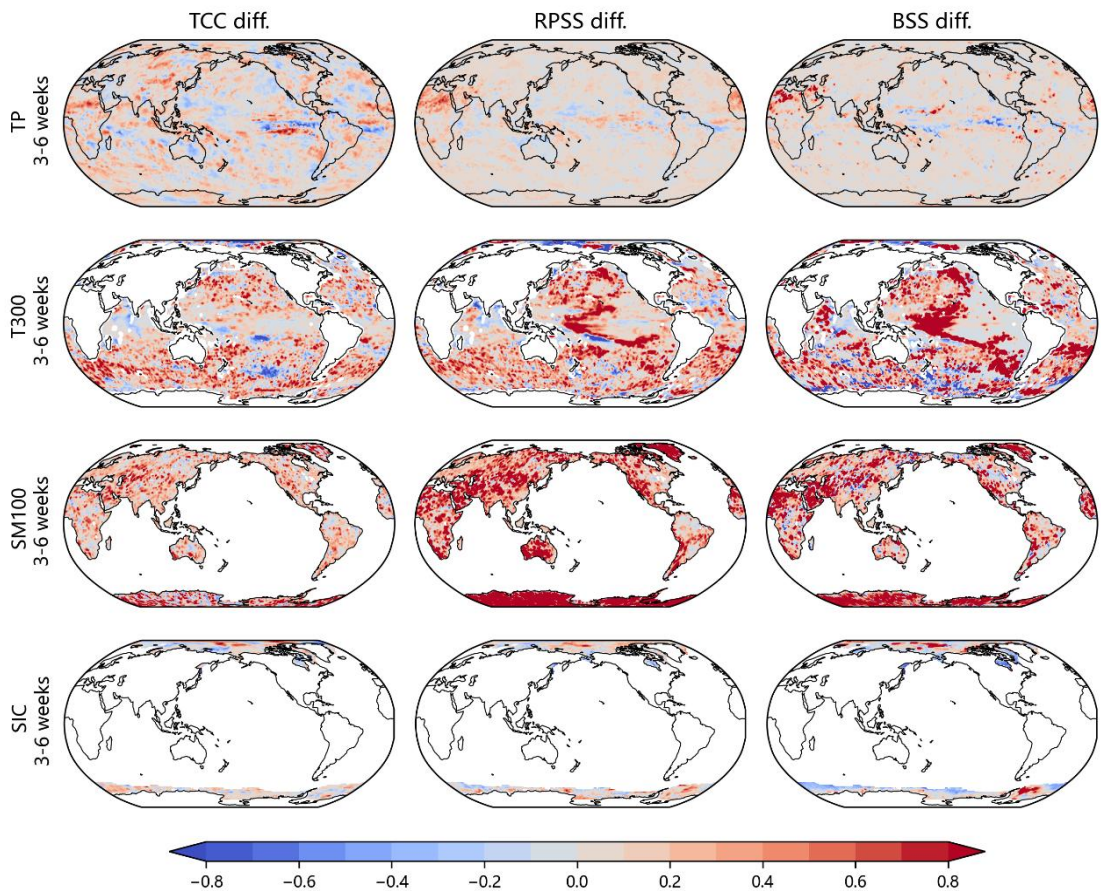

**Figure 2:** Spatial maps of the differences in TCC (first column), Ranked Probability skill Score (RPSS) (second column) and Brier Skill Score (BSS) (third column) between FengShun-CSM forecasts and ECMWF S2S forecasts for TP (first row), T300 (second row), SM100 (third row), and SIC (fourth row) at lead times of 3–6 weeks, using all testing data from 2021.

*Prediction of major intra-seasonal variability modes*

Previous studies have demonstrated that intra-seasonal variability modes, such as the MJO and NAO, serve as critical sources of predictability on sub-seasonal timescales (Hurrell and Deser, 2010; Vitart, 2017; Vitart and Robertson, 2018). The accuracy with which a model predicts these modes directly influences the reliability of sub-seasonal forecasts. Therefore, it is essential to evaluate the forecasting performance of FengShun-CSM for these major intra-seasonal variability modes. This subsection specifically examines the prediction skills of FengShun-CSM for the MJO and NAO, comparing them with the forecast results from ECMWF S2S.

The MJO is widely recognized as the most important source of predictability on sub-seasonal timescales and can influence global climate anomalies through teleconnections (Zhang et al., 2013; Stan et al., 2017; Vitart et al., 2017). In this study, we employed the real-time multivariate MJO (RMM) index (Wheeler and Hendon, 2004) to evaluate the prediction skill of the MJO. The RMM index is derived by projecting observations or model forecasts onto the first two Empirical Orthogonal Functions (EOFs) of OLR and zonal winds at 850 hPa and 200 hPa. Figure 3a displays a comparison of the MJO prediction skills between FengShun-CSM and ECMWF S2S. As the lead time increases, the prediction skills of both models for the MJO gradually decrease. Although this study only assessed the prediction skills for 2021, it is evident that FengShun-CSM significantly outperforms ECMWF S2S in MJO forecasts, with its correlation (COR) coefficient values remaining above 0.7 for up to 42 days. This demonstrates that FengShun-CSM has a strong capability in predicting the MJO.

The NAO is an important extratropical atmospheric variability mode that significantly impacts weather and climate across the Northern Hemisphere, particularly in Europe and eastern North America (Hurrell et al., 1995). This influence is typically more pronounced during the Northern Hemisphere winter (Nie et al., 2019). In this study, due to the inability of the 2021 testing data to cover a complete winter season (from December to February of the following year), we focused on analyzing daily NAO index predictions initialized from October to December 2021. Figure 3b presents a comparison of the NAO prediction skills between FengShun-CSM and ECMWF S2S. With only 26 initialization days available during this period, the limited sample size resulted in instability in the NAO prediction skills of both models. Overall, FengShun-CSM exhibits comparable skill to ECMWF S2S. When applying a COR threshold of 0.5 to determine skillful NAO forecasts, ECMWF S2S achieves a prediction skill of only 13 days, while FengShun-CSM extends this to 15 days, indicating that FengShun-CSM is more proficient in predicting the NAO than ECMWF S2S.

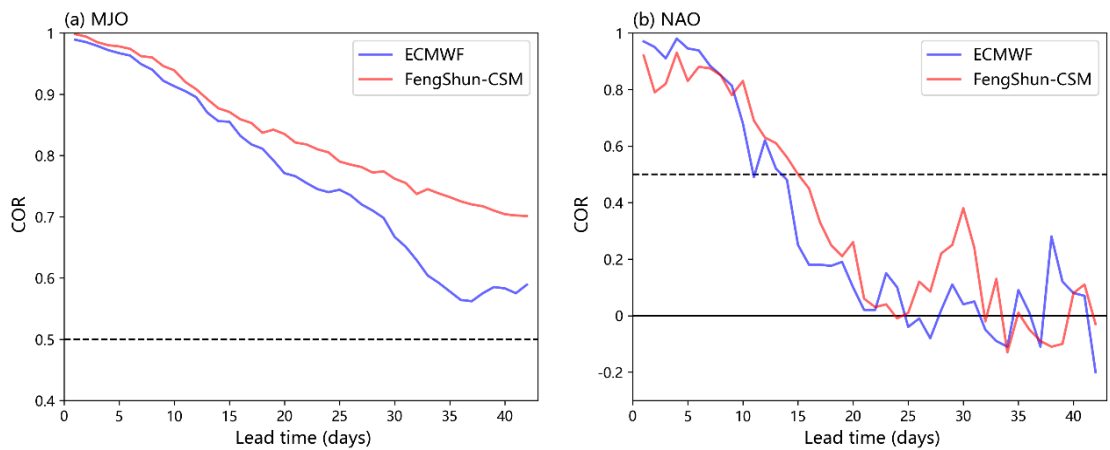

**Figure 3:** Comparison of the correlation (COR) skills for (a) the real-time multivariate Madden–Julian Oscillation (MJO) (RMM) index and (b) the North Atlantic Oscillation (NAO) index between ECMWF S2S forecasts (in blue) and FengShun-CSM forecasts (in red). The COR values for the RMM and NAO indices are plotted as a function of forecast lead time. The dashed black line represents the prediction skill threshold of COR = 0.5.

*Prediction of extreme events*

In mid-February 2021, central North America experienced an extreme cold wave event, causing significant disaster impacts. In Texas alone, over 1000 power generators malfunctioned, leaving nearly 10 million people without electricity and water for several weeks and resulting in at least 210 deaths. This event became the costliest winter storm ever record in the United States (NCEI, 2022). Subsequently, during the summer of 2021, a widespread heatwave struck central and western North America from late June to early July. Several cities broke historical high-temperature records, with regional maximum daily average temperatures exceeding the climatological mean by approximately four standard deviations. This led to severe drought conditions and substantial economic losses (Map Archive, 2021; Thompson et al., 2022). Therefore, it is important to evaluate the sub-seasonal forecast capability for such extreme events.

Figure 4a shows a comparison of the standardized minimum 2 m temperature (T2M_MIN) anomalies among the observations from ERA5, ECMWF S2S, and FengShun-CSM, averaged across central North America by 91°W to 106°W in longitude and 28°N to 44°N in latitude. Observations are temporally averaged over ten days from February 10th to February 19th. Forecasts from FengShun-CSM and ECMWF S2S were initialized on different dates. Notably, ECMWF S2S predicts positive T2M_MIN anomalies for forecasts initialized before January 25th. Although ECMWF S2S capture negative T2M_MIN anomalies for forecasts initialized after January 25th, the model consistently underestimates the intensity of T2M_MIN. In contrast, FengShun-CSM predicts negative anomalies across all lead times, providing a lead time of one month prior to the occurrence of the event. Furthermore, the spatial distributions of standardized T2M_MIN anomalies show that patterns predicted by FengShun-CSM aligns more closely with observations (Figure 4b), which is crucial for meteorological disaster prevention and mitigation. In summary, compared to ECMWF S2S, FengShun-CSM demonstrates superior performance in predicting both the intensity and spatial distribution of the cold wave event, with a longer lead time.

The prediction skills for the 2021 summer heatwave event that occurred in the central and western North America are shown in Figures 4c and 4d. The standardized anomalies in Figure 4a were calculated using the same method as for the cold wave event, but based on maximum 2 m temperature (T2M_MAX) data from June 25th to July 2nd, with spatially averaged over the central and western North America (107°W to 130°W in longitude and 49°N to 60°N in latitude). The standardized anomaly for observed T2M_MAX is approximately four standard deviations above the climatological mean. Although both FengShun-CSM and ECMWF S2S captured the occurrence of the event across most lead times, they significantly underestimated its intensity and even predicted negative T2M_MAX anomalies at certain lead times. Overall, FengShun-CSM demonstrates comparable skill to ECMWF S2S but predicts higher intensity and a more reasonable spatial distribution for forecasts initialized on June 7th and May 27th, outperforming ECMWF S2S.

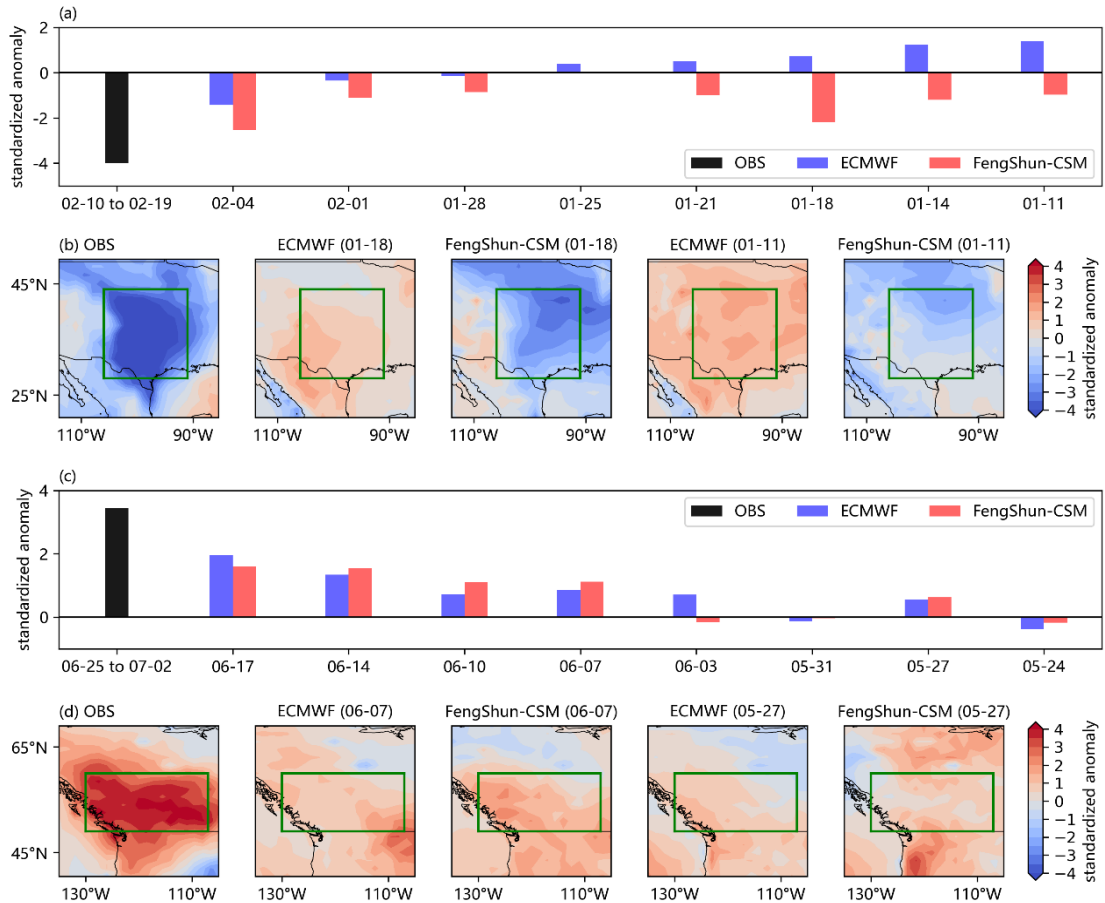

**Figure 4:** Comparison of (a) spatially and temporally averaged standardized minimum 2 m temperature (T2M_MIN) anomaly from 10–19 February 2021. ERA5 observation (in black) is compared with ECMWF S2S forecasts (in blue) and FengShun-CSM forecasts (in red), initialized on: 4 February, 1 February, 28 January, 25 January, 21 January, 18 January, 14 January, and 11 January. Comparison of (b) temporally averaged standardized T2M_MIN anomaly maps for ERA5 observation (first column), ECMWF S2S initialized on 18 January (second column), FengShun-CSM initialized on 18 January (third column), ECMWF S2S initialized on 11 January (fourth column), and FengShun-CSM initialized on 11 January (fifth column). The green contour highlights the central region of the cold wave event that occurred in central North

America in February 2021. (c) Same as (a), but for spatially and temporally averaged standardized maximum 2 m temperature (T2M_MAX) anomaly from 25 June to 2 July 2021. ECMWF S2S and FengShun-CSM forecasts are both initialized on: 17 June, 14 June, 10 June, 7 June, 3 June, 31 May, 27 May, and 24 May. (d) Same as (b), but for temporally averaged standardized T2M_MAX anomaly maps for ERA5 observation (first column), ECMWF S2S initialized on 7 June (second column), FengShun-CSM initialized on 7 June (third column), ECMWF S2S initialized on 27 May (fourth column), and FengShun-CSM initialized on 27 May (fifth column). The green contour highlights the central region of the Summer 2021 heatwave event in central and western North America.

Supplementary Figure 9 compares the performance of FengShun-CSM and ECMWF S2S in forecasting an extreme precipitation event that occurred in central and western Europe in mid-July 2021 (Tuel et al., 2022). The spatial distribution of standardized TP anomalies shows that ECMWF S2S predicted negative TP anomalies. In contrast, FengShun-CSM demonstrates better forecast performance in terms of both the intensity and spatial distribution of extreme precipitation. However, it is important to note that, compared to observations, the intensity predicted by FengShun-CSM is slightly weaker, and there is a certain deviation in the location of the extreme precipitation center. This suggests that there is still room for further improvement and refinement of the model.

This study highlights the proficiency of FengShun-CSM in predicting extreme events. We attribute its superior performance to the model's integration of multi-sphere information, particularly the influences of the upper ocean, land, and sea ice. These factors have been demonstrated to be important sources for improving sub-seasonal prediction skills (Koster et al., 2010; Jeong et al., 2013; Vitart, 2017). Exploring how these factors affect the model's prediction performance is crucial for identifying key sources of predictability and further enhancing sub-seasonal forecast accuracy. However, such analysis extends beyond the scope of the current study and will be explored in future research.

*Inter-sphere coupling relationships in FengShun-CSM model*

With the parameterization of physical processes in dynamical models continues to improve, the simulation capabilities of dynamical models are steadily enhancing. Meanwhile, the coupling relationships between different spheres in dynamical models adhere well to physical conservation laws (Zhou et al., 2020; Wu et al., 2021). As the first AI-based climate system model, FengShun-CSM's evaluation of its predictive performance in inter-sphere coupling relationships is of great significance for continuously refining the coupling mechanisms of models and further enhancing sub-seasonal prediction skills. Furthermore, through such evaluations, we can deepen our understanding of complex interactions in climate systems while providing guidance for subsequent technological advancements.

Figure 5 illustrates the forecasting performance of FengShun-CSM at 5–6 weeks lead for global atmosphere-ocean, atmosphere-land, and atmosphere-ice coupling relationships diagnostics. When the patterns predicted by the model are closer to the observed patterns, it indicates that the model is proficient in predicting the relationships between different spheres. The atmosphere-land coupling plays an important role in the dynamics of the hydrologic cycle, and the strong feedback between soil moisture and precipitation has been widely studied (Eltahir and Pal, 1996; Giorgi et al., 1996; Findell and Eltahir, 1997). In this study, we quantify CORs between soil moisture and precipitation, revealing FengShun-CSM's superior capability in replicating observed spatial covariation patterns (Figure 5). This is further corroborated by robust SM100-T2m and SM100-MSL relationships (Supplementary Figure 11), demonstrating enhanced land-atmosphere interaction emulation. For atmosphere-ocean coupling relationship forecasts, we calculated the CORs between SSH and MSL (Figure 5), as well as between MLT and MSL (Supplementary Figure 12). Although FengShun-CSM can roughly predict the spatial distribution patterns of atmosphere-ocean coupling relationships, its prediction biases are larger compared to ECMWF S2S. This indicates that FengShun-CSM's ability to simulate atmosphere-ocean coupling relationships still requires further improvement and enhancement. Additionally, regarding atmosphere-ice coupling relationships, previous studies have shown that there is a complex interaction between SIC and T2m, primarily characterized by a two-way feedback mechanism (Screen and Simmonds, 2010; Stroeve et al., 2012). Here we calculated the CORs between SIC and T2m (Figure 5). The results show that both FengShun-CSM and ECMWF S2S are able to capture the atmosphere-ice coupling relationships in certain regions, with their performances being roughly comparable. Notably, both models perform better in the Antarctic region. However, considering the demand for more accurate sub-seasonal predictions in Arctic shipping, FengShun-CSM's simulation of atmosphere-ice coupling in the Arctic region still requires further improvement. It is important to note that this study only analyzes concurrent relationships, the lead-lag influences will be explored in depth in future research.

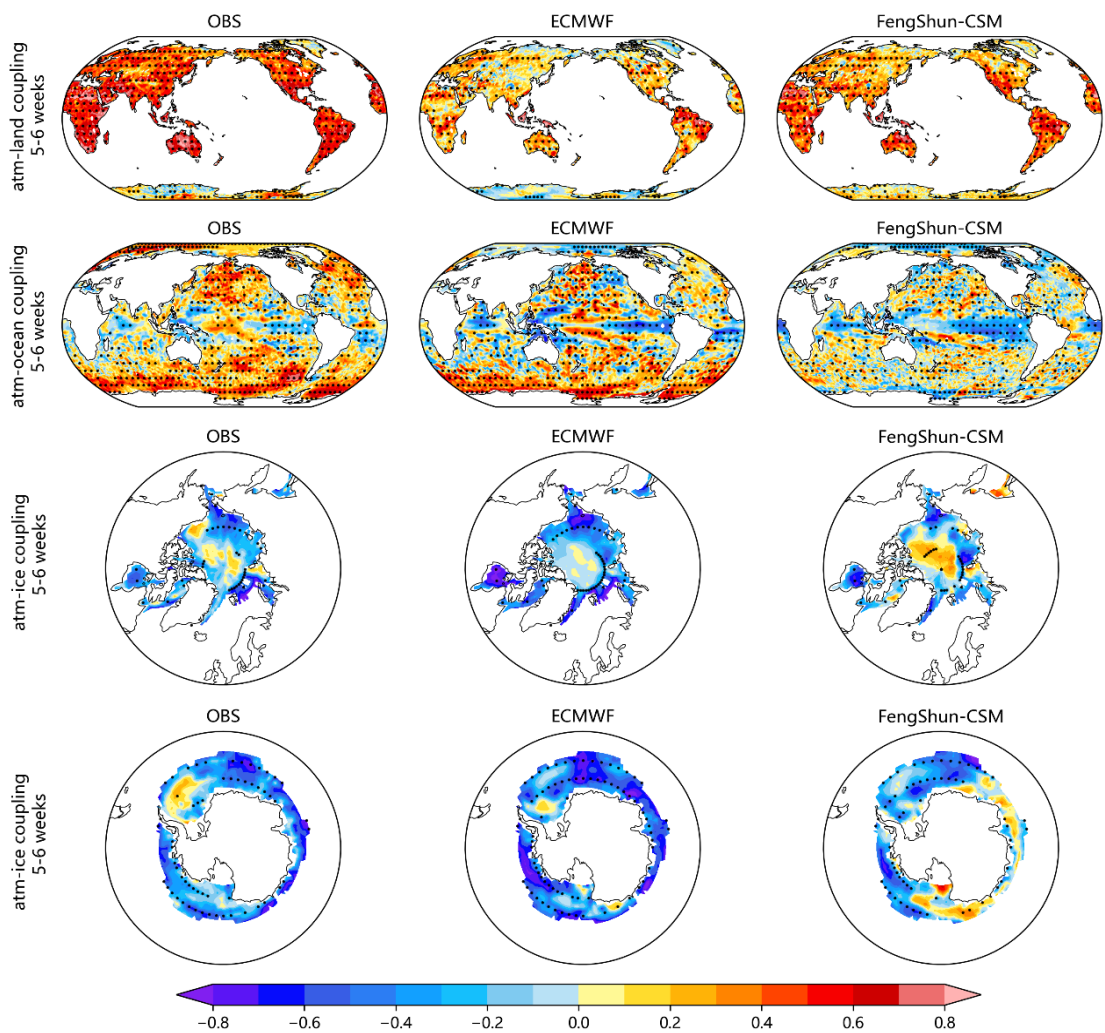

**Figure 5:** Spatial maps of atmosphere-land coupling (first row), atmosphere-ocean

coupling (second row), atmosphere-ice coupling over the Arctic region (third row), and atmosphere-ice coupling over the Antarctic region (fourth row) for OBS (first column), ECMWF S2S forecasts (second column), and FengShun-CSM forecasts (third column) at lead times of 5–6 weeks. The analysis is based on all testing data from 2021. In each subplot, the color scale represents the COR between different Earth system components. Black dotted areas indicate regions where the correlations are statistically significant at the 95% confidence level.

**Discussions**

In this study, we built FengShun-CSM, an AI-based climate system model that achieves multi-sphere coupling of the atmosphere, ocean, land, and sea ice through machine learning technology. This model provides global forecasts of daily mean values for up to 60 days, with a daily temporal resolution and a spatial resolution of 0.25°, including 17 atmospheric variables, 4 land variables, 6 oceanic variables, and 2 sea ice variables. The performance of FengShun-CSM was evaluated against ERA5 and Ocean Reanalysis System 5 (ORAS5) reanalysis datasets and compared with ECMWF S2S forecasts, using all testing data from 2021. The results demonstrate that FengShun-CSM outperforms ECMWF S2S in predicting most variables, whether in deterministic, probabilistic, or even extreme forecasts. Considering the significant influence of intra-seasonal climate variability modes on sub-seasonal forecast skill, accurately predicting these modes is particularly crucial. FengShun-CSM exhibits strong performance in predicting the MJO, maintaining a prediction skill exceeding 0.7 up to 42 days in advance. Moreover, FengShun-CSM extends the skillful NAO forecasts from 13 days to 15 days. This capability may be a key factor contributing to FengShun-CSM's superior performance in predicting most variables. Furthermore, FengShun-CSM has shown huge potential in predicting extreme events on sub-seasonal timescales, such as cold waves, heatwaves, and floods. This feature highlights the practical value of FengShun-CSM, as its early disaster warning information is of great significance for disaster prevention and mitigation efforts.

In addition to its exceptional accuracy, FengShun-CSM exhibits a distinguishing advantage in computational efficiency. Traditional physics-based climate models are facing increasing computational demands due to increasingly sophisticated representations of physical processes, while AI-driven FengShun-CSM achieves rapid large-ensemble forecasting with remarkable efficiency. This computational advantage holds critical importance for time-sensitive disaster preparedness, as it enables near-real-time generation of actionable forecasts. Furthermore, the model's high computational efficiency creates opportunities to enhance sub-seasonal forecasting capabilities through ensemble expansion—a proven approach for reducing forecast uncertainty (Chen et al., 2024). Notably, FengShun-CSM achieves this scalability potential while maintaining relatively modest increases in computational resource requirements, offering a practical pathway to substantially enhance prediction reliability without prohibitive infrastructure investments.

While demonstrating notable capabilities and high computational efficiency, the FengShun-CSM model exhibits critical structural limitations that constrain its Earth system modeling. The model's vertically limited atmospheric representation (13 pressure levels with a 50 hPa upper boundary) precludes robust simulation of stratospheric dynamics, notably omitting key phenomena such as the Quasi-Biennial Oscillation (QBO) and Sudden Stratospheric Warmings (SSW). Furthermore, while the model integrates the core spheres of the Earth system (atmosphere, ocean, land, and sea ice), it lacks critical biogeochemical processes including vegetation variation, ocean wave interactions, carbon cycle feedbacks, and ecosystem responses—elements increasingly recognized as essential for advancing sub-seasonal predictability (Baldwin et al., 2003; Collins et al., 2017; Song et al., 2020; Lian et al., 2022). These mechanistic gaps hinder its evolution into a comprehensive machine learning-based Earth System Model (ESM). Additionally, the real-time forecasts of ECMWF S2S model (cycle C47r2) consist of 51 ensemble members, and the latest version has expanded to 101 members, while the current FengShun-CSM only includes 48 members. Although FengShun-CSM demonstrates superior prediction accuracy through initial condition perturbation strategies in its atmospheric and oceanic components (as detailed in "FengShun-CSM model"), its capacity to resolve critical coupling mechanisms— particularly atmosphere-ocean interactions and atmosphere-ice linkages—remains limited. This limitation becomes particularly evident when addressing S2S and longer-term climate predictability, where accurate representation of cross-component feedbacks is paramount for forecast reliability.

Therefore, the future directions for improving FengShun-CSM include increasing the number of ensemble members from 48 to 100, extending its atmospheric top level beyond 50 hPa, incorporating additional key physical processes, and adopting more efficient ensemble forecasting strategies to enhance prediction accuracy. These enhancements will help improve the overall performance of the model and lay the foundation for achieving reliable climate predictions on longer timescales.

**Methods**

*Data*

In this study, we utilized reanalysis data from ERA5 and ORAS5 to train and test the FengShun-CSM model, as well as evaluate its sub-seasonal forecasting performance. Both the ERA5 and ORAS5 datasets feature a spatial resolution of 0.25° and a temporal resolution of one day. Specifically, atmospheric and land variables in the training and test datasets were obtained from ERA5, while oceanic and sea ice variables were sourced from ORAS5.

FengShun-CSM's training relied on 28 years of data spanning from 1993 to 2020, while evaluation was conducted using testing data from 2021. The FengShun-CSM is a comprehensive climate system model which is capable of forecasting 29 variables, including 5 upper-air atmospheric variables across 13 pressure levels (50, 100, 150, 200, 250, 300, 400, 500, 600, 700, 850, 925, and 1000 hPa), 12 single-air variables, 4 land variables, 4 upper-ocean variables across 13 depth levels (0.50, 2.66, 5.14, 8.09, 11.77, 16.52, 22.75, 30.87, 41.18, 53.85, 69.02, 86.92, 108.0, 133.07, 163.16, 199.79, 221.14, 271.35, 333.86, 411.79, 508.63 meters), 2 single-ocean variables, and 2 sea ice variables. Among the upper-air atmospheric variables are geopotential (Z), temperature (T), u component of wind (U), v component of wind (V), and specific humidity (Q). The single-air variables include T2M, T2M_MIN, T2M_MAX, 2 m dewpoint temperature (D2M), 10 m u wind component (U10), 10 m v wind component (V10), 100 m u wind component (U100), 100 m v wind component (V100), mean sea-level pressure (MSL), total cloud cover (TCC), OLR, and TP. The land variables consist of soil moisture from 0 to 289 cm (SM), soil temperature from 0 to 289 cm (ST), snow density (RSN), and snow depth (SD). The upper-ocean variables include potential temperature (PT) and salinity (S), u component of surface current (OCU), and u component of surface current (OCV). The two single-ocean variables are SSH and MLT. Lastly, the sea ice variables are SIC and SIT. Supplementary Table 1 provides a detailed list of these variables along with their abbreviations.

In our research, we employed ECMWF S2S forecasts generated from model cycle C47r2 to compare the forecasting performance of FengShun-CSM. Specifically, the ECMWF S2S reforecasts are initialized twice weekly, consisting of 10 ensemble members, and span initialization dates from 2006 to 2020. For the real-time forecast in 2021, the initialization dates remained consistent with those of reforecasts, but the number of ensemble members was increased to 51. Anomalies for all variables were defined as deviations from the climatological mean, which was calculated over the period from 2006 to 2020. Furthermore, we generated a set of hindcasts for FengShun-CSM spanning 2006 to 2020 to establish its climatology. The anomalies for the 2021 FengShun-CSM forecasts were then computed by subtracting this climatology. It is worth noting that the spatial resolution of atmospheric and land variables in ECMWF S2S forecasts is 1.5°, while the resolution for oceanic and sea ice variables is 1°. To ensure equitable comparisons, all datasets were uniformly interpolated to a spatial resolution of 1.5°. This approach enables a direct and fair assessment between FengShun-CSM and ECMWF S2S.

*FengShun-CSM model*

FengShun-CSM is a global sub-seasonal forecasting AI model that achieves fully coupled atmosphere-ocean-land-sea ice interactions using machine learning technology. As illustrated in Figure 6, FengShun-CSM model includes three key sub-modules: the atmospheric prediction module, the ocean prediction module, and the coupling module.

Specifically, the atmospheric prediction module integrates atmospheric and land variables, and the ocean prediction module incorporates oceanic and sea ice variables. Both modules have the same structure, which is an ensemble forecast model based on initial condition perturbations. By inputting atmospheric/land variables ($A_{t-24h}$ and $A_t$) and oceanic/sea ice variables ($O_{t-24h}$ and $O_t$) from two consecutive days, the model predicts the atmospheric/land variables ($A_{t+24h}$) and oceanic/sea ice variables ($O_{t+24h}$) for the following day in an autoregressive manner, with each time step representing a 24-hour interval.

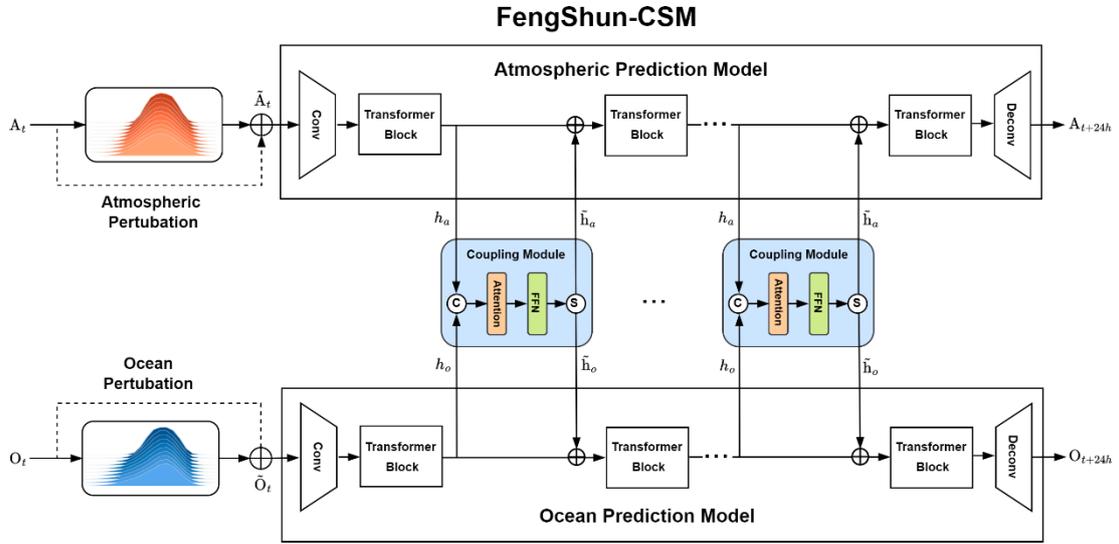

**Figure 6:** Schematic diagram of the structures of FengShun-CSM model. The atmospheric module integrates elements of the atmosphere and land, and the ocean module incorporates elements of the ocean and sea ice. The two modules exchange information and influence each other through coupling modules.

The ensemble forecast model based on initial condition perturbations consists of two modules: the perturbation module and the prediction module. The perturbation module transforms input data ($X$) into a Gaussian distribution, which captures the probabilistic characteristics of input data, and then samples a perturbation vector ($Z$) from the Gaussian distribution. The perturbation vector has the same dimension as the input data. The final ensemble output ($\hat{X} = X + Z$) is generated and used as the input for the prediction module to predict the atmospheric/land and oceanic/sea ice variables for the following day.

The coupling module serves as a bridge connecting the atmospheric ensemble module and the ocean ensemble module to facilitate information exchange. Specifically, the coupling module is embedded between multiple transformer blocks in both atmospheric and ocean ensemble modules, with one coupling block introduced every four transformer blocks. The coupling block is implemented using a transformer block, which consists of an attention layer and a feedforward neural network (FFN) layer. However, it differs from the transformer blocks in the atmospheric and ocean ensemble modules in that its input is obtained by concatenating atmospheric and oceanic features ($h_a$ and $h_o$), thus doubling the feature dimension. After the concatenated features are updated, they are split into atmospheric and oceanic features ($\tilde{h}_a$ and $\tilde{h}_o$), which are then added back to their respective main features. To ensure that the atmospheric and ocean ensemble modules operate independently without interference during the early stages of training, we initialize the residual branch of the coupling module to zero. As training progresses, the output of the coupling module gradually integrates the features resulting from the interactions between the atmospheric ensemble module and the ocean ensemble module, and then influences them separately. In other words, features from the atmospheric ensemble module affect the ocean ensemble module, and features from the ocean ensemble module also influences the atmospheric ensemble module. The location and intensity of these influences are learned and implemented through the attention mechanism.

*Evaluation method*

In our research, each variable in the 6-week (42-day) forecast undergoes a detrending process to eliminate linear trends before evaluation. Specifically, a linear regression model is established to estimate the weekly mean linear trend of both predictions and observations during the hindcast period (2006–2020). For the testing period (2021), this model uses the weekly data from 2021 as input to calculate the trend values, which are then subtracted from both the predictions and observations to obtain the detrended fields. Subsequently, the TCC is used to evaluate the deterministic metrics

of the ensemble mean. TCC is calculated as follows:

$$TCC_{i,t} = \frac{\sum_{j=1}^{N}(P_{j,t}-\overline{P_t})(O_{j,t}-\overline{O_t})}{\sqrt{\sum_{j=1}^{N}(P_{j,t}-\overline{P_t})^2}\sqrt{\sum_{j=1}^{N}(O_{j,t}-\overline{O_t})^2}} \quad (1)$$

where $i$ denotes the grid points (latitude and longitude coordinates). $t$ refer to lead time steps. $j$ represents the initialization time in the testing dataset. N is equal to 104, which refers to 104 initialization time in 2021. $P$ and $O$ are the anomaly of the prediction and observation, respectively. $\overline{P}$ and $\overline{O}$ represent the average of the prediction and observation for all initialization times in 2021, respectively.

To evaluate the performance of the ensemble forecast, we utilize the RPSS (Epstein 1969), which quantifies the comparison between the cumulative squared probability errors of a given forecast and those of a climatological forecast. The calculation of the RPSS metric requires prior determination of the ranked probability scores (RPS) for both the model's forecast ($RPS_{forecast}$) and the climatological forecast ($RPS_{clim}$). $RPS_{forecast}$ and $RPS_{clim}$ are calculated as follows:

$$RPS_{forecast} = \sum_{k=1}^{K}(\sum_{i=1}^{k}p_{forecast}(i) - \sum_{i=1}^{k}p_o(i))^2 \quad (2)$$

$$RPS_{clim} = \sum_{k=1}^{K}(\sum_{i=1}^{k}p_{clim}(i) - \sum_{i=1}^{k}p_o(i))^2 \quad (3)$$

where $K$ denotes the number of categories, and in this study, $K$ equals 3. $p_{forecast}(i)$, $p_{clim}(i)$, and $p_o(i)$ represent forecasted, climatological, and observed probability of the event's occurrence in category $i$.

The RPS value ranges between 0 and 1, where a lower value indicates smaller prediction probability errors, thus reflecting a more accurate forecast. Based on the RPS values of the forecast and the climatological forecast, the RPSS can be determined as:

$$RPSS = 1 - \frac{\langle RPS_{forecast}\rangle}{\langle RPS_{clim}\rangle} \quad (4)$$

where the brackets $\langle ... \rangle$ denote the average of the $RPS_{forecast}$ and $RPS_{clim}$ values across all forecast-observation pairs. The RPSS value ranges from $-\infty$ to 1, where a value of 1 corresponds to a perfect forecast, and higher values indicate better forecasting performance.

Additionally, we employ the BSS (Wilks, 2011) to evaluate the performance of extreme forecasts. The BSS can be regarded as a special case of the RPSS for binary forecasts. In this study, we use the 90th climatological percentile as the threshold for extreme events. The BSS is calculated as follows:

$$BSS = 1 - \frac{\langle BS_{forecast} \rangle}{\langle BS_{clim} \rangle} \qquad (5)$$

where $BS_{forecast}$ and $BS_{clim}$ represent the Brier Score (BS) for the model's forecast and the climatological forecast, respectively. Similar to the range of RPSS values, a BSS value of 1 indicates a perfect forecast, and smaller values indicate poorer forecasting performance.

**Acknowledgements**

This work was supported by the National Key R&D Program of China under Grant 2021YFA0718000.

**Data availability**

We downloaded a subset of the daily atmospheric and land's statistics on single and pressure levels from the ERA5 daily data from the official website of Copernicus Climate Data (CDS) at https://cds.climate.copernicus.eu/datasets/derived-era5-single-levels-daily-statistics?tab=overview and https://cds.climate.copernicus.eu/datasets/derived-era5-pressure-levels-daily-statistics?tab=overview. The daily statistics for the ocean and sea-ice that we downloaded are sourced from the ORAS5 daily data from the official website of Copernicus Marine Data at https://data.marine.copernicus.eu/product/GLOBAL_MULTIYEAR_PHY_ENS_001_031/description. The ECMWF S2S data were obtained from https://apps.ecmwf.int/datasets/data/s2s/.

**Supplementary Information**

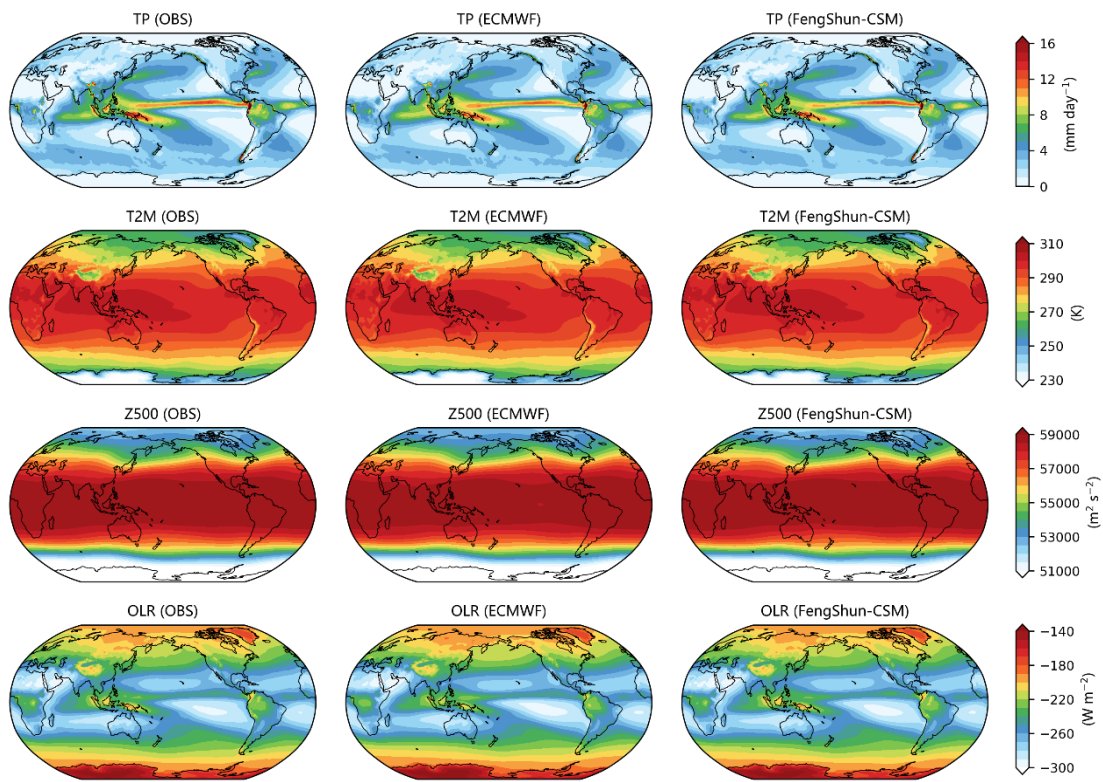

**Supplementary Figure 1:** Climatological spatial maps of TP (first row), T2M (second row), Z500 (third row), and OLR (fourth row) for OBS (first column), ECMWF S2S forecasts (second column), and FengShun-CSM forecasts (third column) at lead times of 3–6 weeks, based on all testing data from 2021.

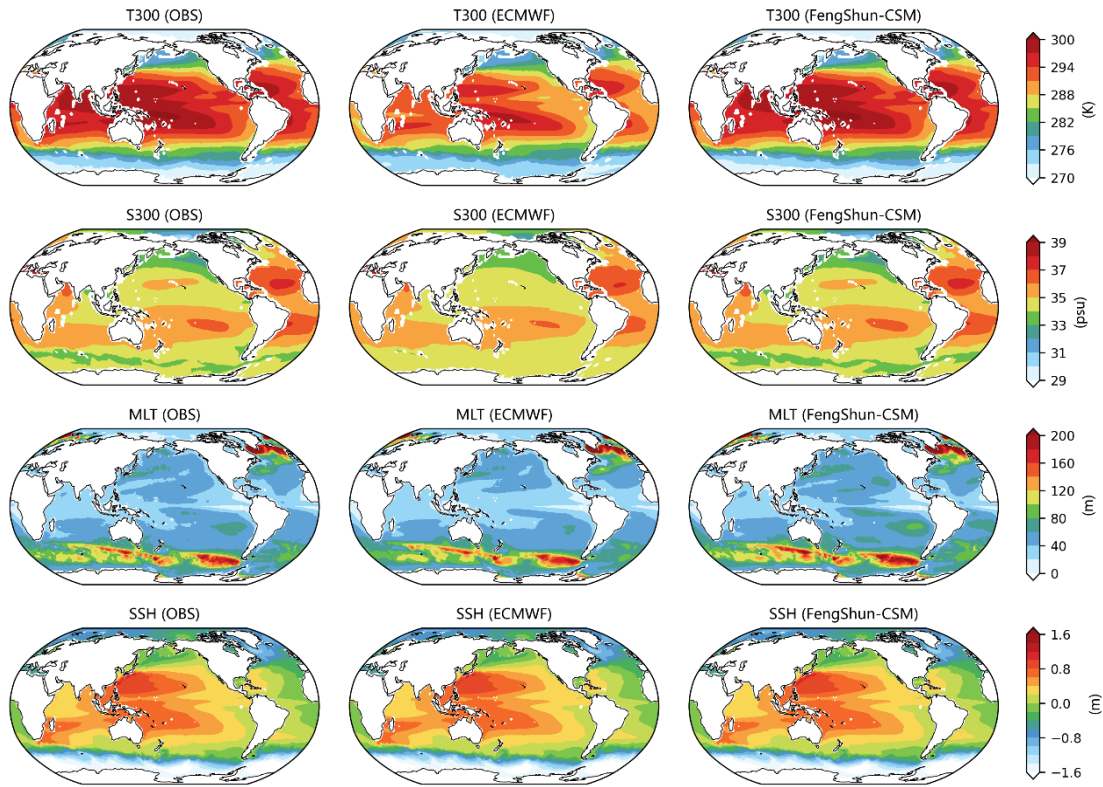

**Supplementary Figure 2:** Climatological spatial maps of T300 (first row), S300 (second row), MLT (third row), and SSH (fourth row) for OBS (first column), ECMWF S2S forecasts (second column), and FengShun-CSM forecasts (third column) at lead times of 3–6 weeks, based on all testing data from 2021.

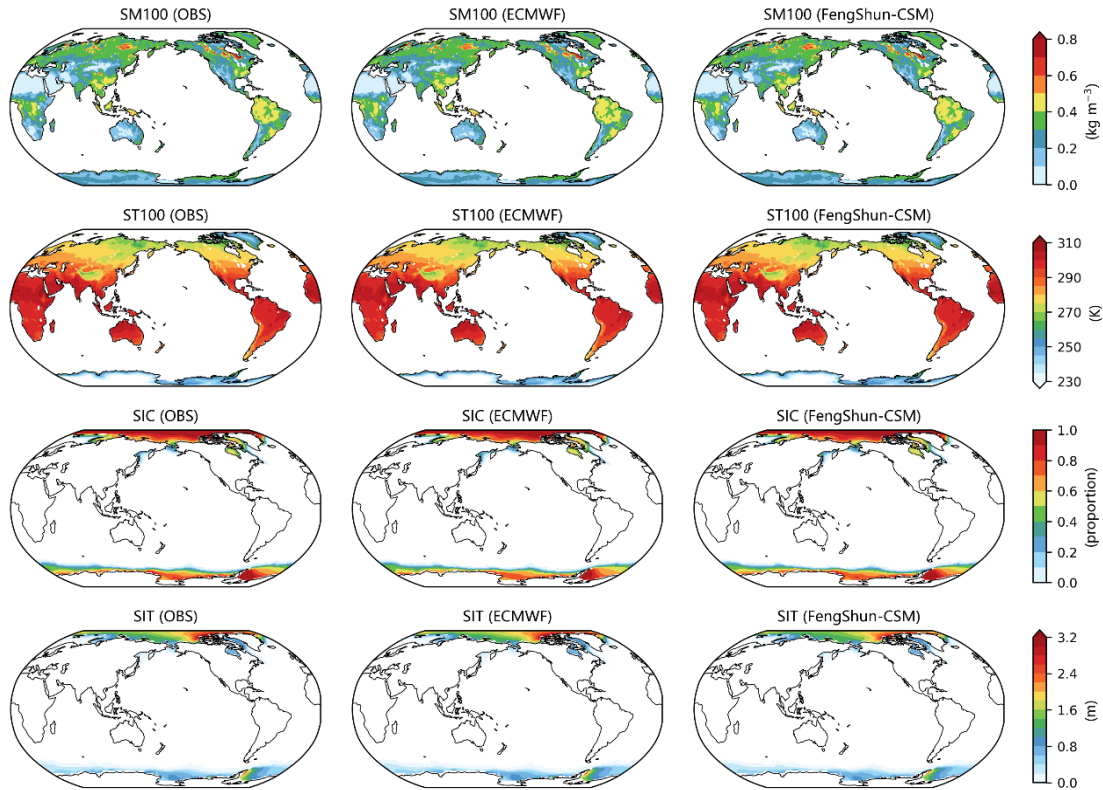

**Supplementary Figure 3:** Climatological spatial maps of SM100 (first row), ST100 (second row), SIC (third row), and SIT (fourth row) for OBS (first column), ECMWF S2S forecasts (second column), and FengShun-CSM forecasts (third column) at lead times of 3–6 weeks, based on all testing data from 2021.

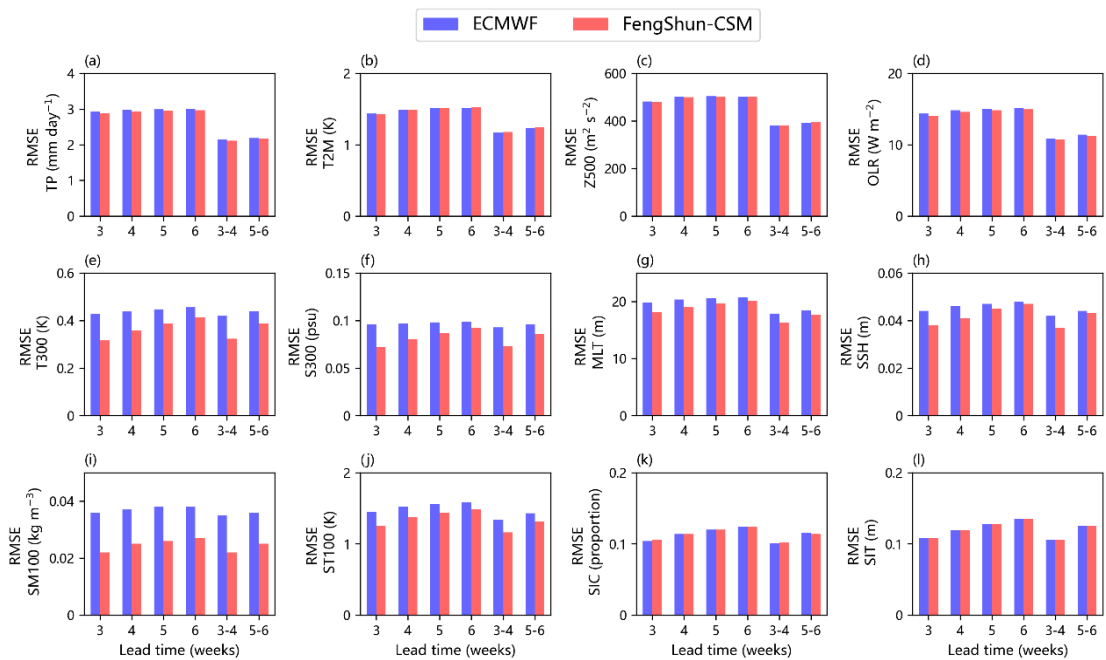

**Supplementary Figure 4:** Comparison of globally-averaged and latitude-weighted

RMSE of the ensemble mean between ECMWF S2S forecasts (in blue) and FengShun-CSM forecasts (in red) for TP, T2M, Z500, OLR, T300, S300, MLT, SSH, SM100, ST100, SIC, and SIT, using all testing data from 2021.

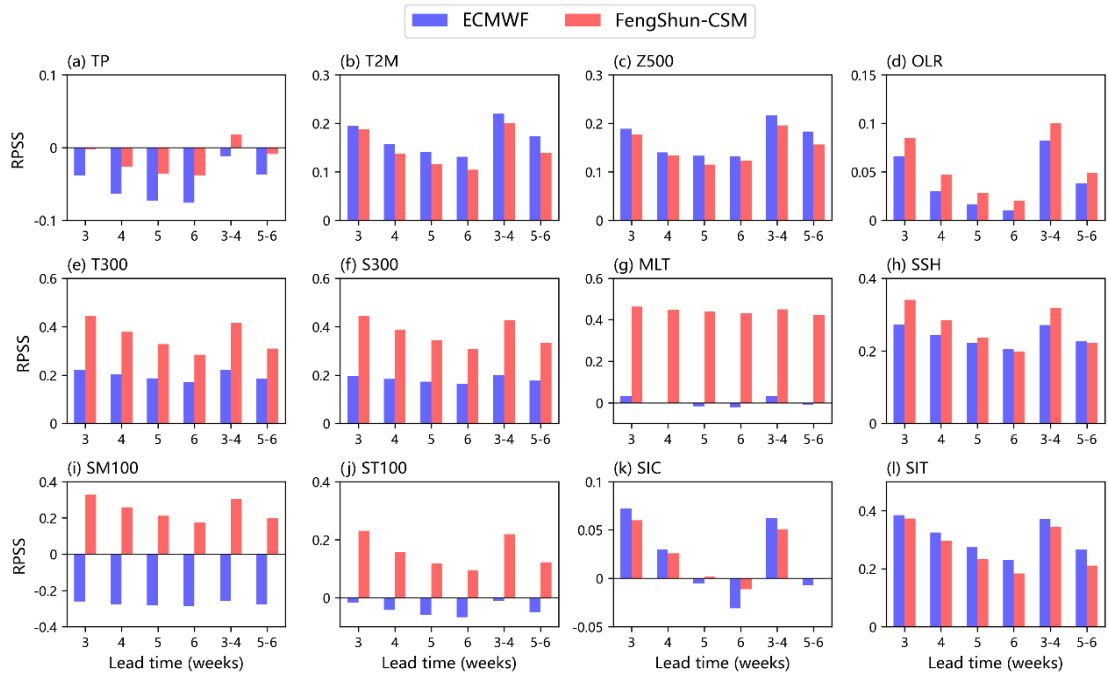

**Supplementary Figure 5:** Comparison of globally-averaged and latitude-weighted RPSS of ECMWF S2S forecasts (in blue) and FengShun-CSM forecasts (in red) for TP, T2M, Z500, OLR, T300, S300, MLT, SSH, SM100, ST100, SIC, and SIT, using all testing data from 2021.

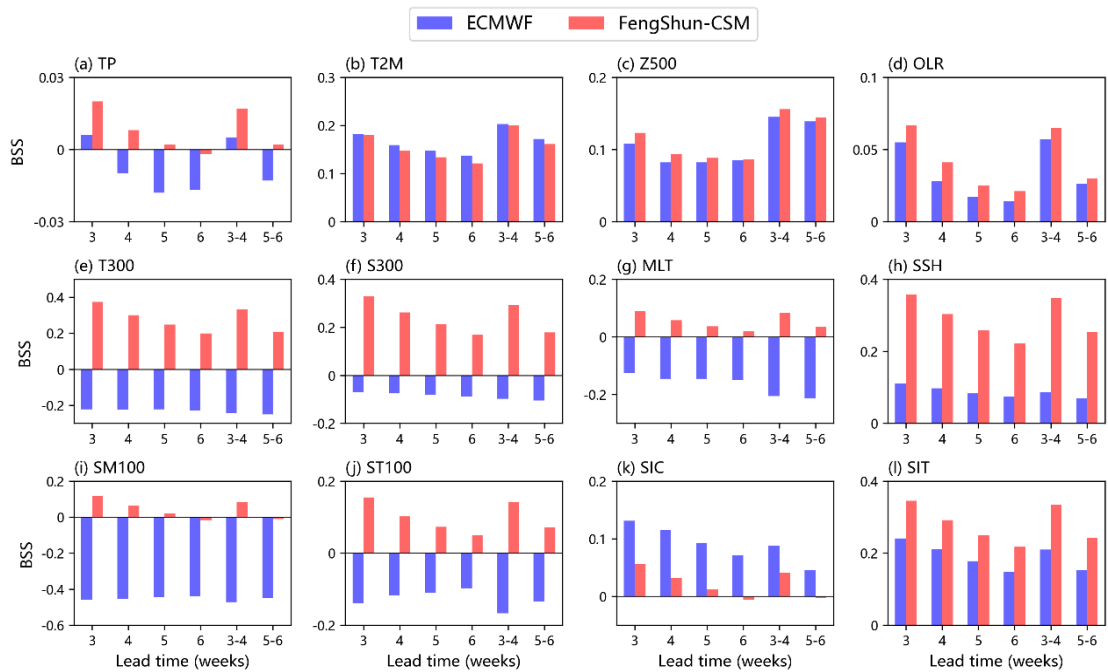

**Supplementary Figure 6:** Comparison of globally-averaged and latitude-weighted BSS of ECMWF S2S forecasts (in blue) and FengShun-CSM forecasts (in red) for TP, T2M, Z500, OLR, T300, S300, MLT, SSH, SM100, ST100, SIC, and SIT, using all testing data from 2021.

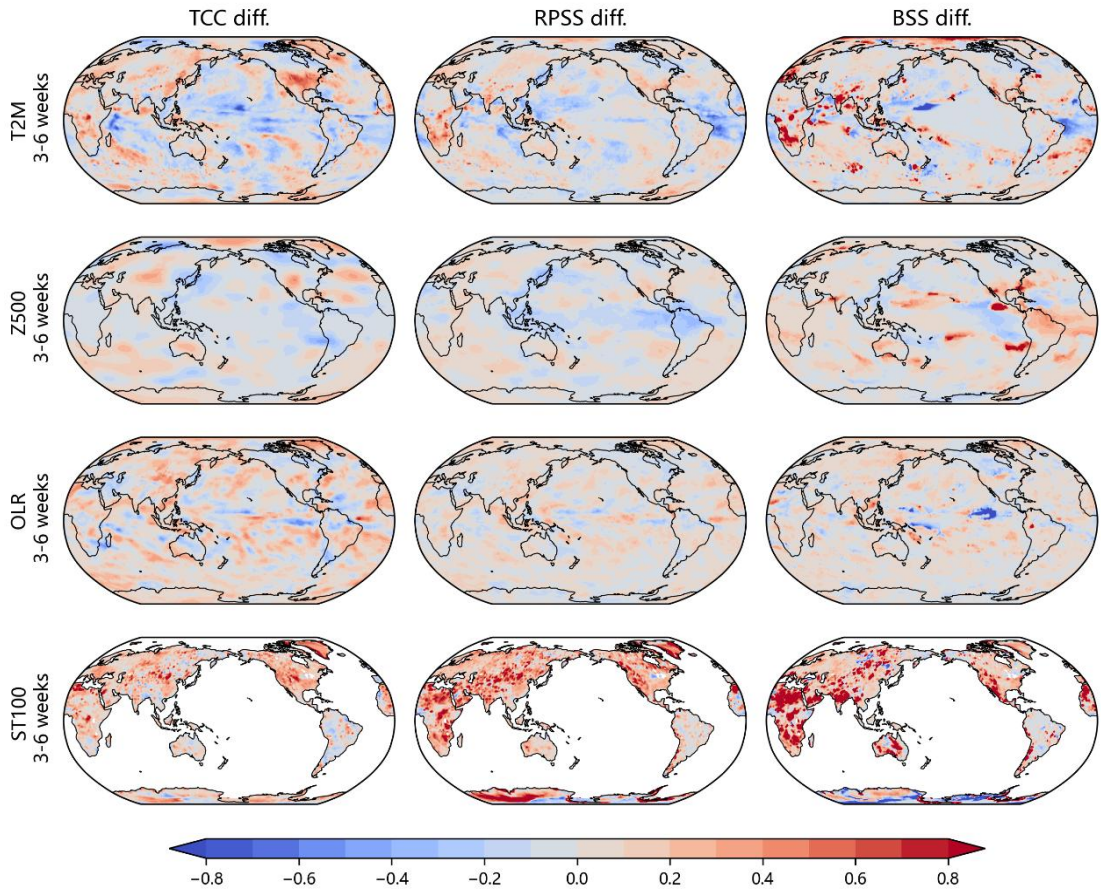

**Supplementary Figure 7:** Spatial maps of the differences in TCC (first column), RPSS (second column) and BSS (third column) between FengShun-CSM forecasts and ECMWF S2S forecasts for T2M (first row), Z500 (second row), OLR (third row), and ST100 (fourth row) at lead times of 3–6 weeks, using all testing data from 2021.

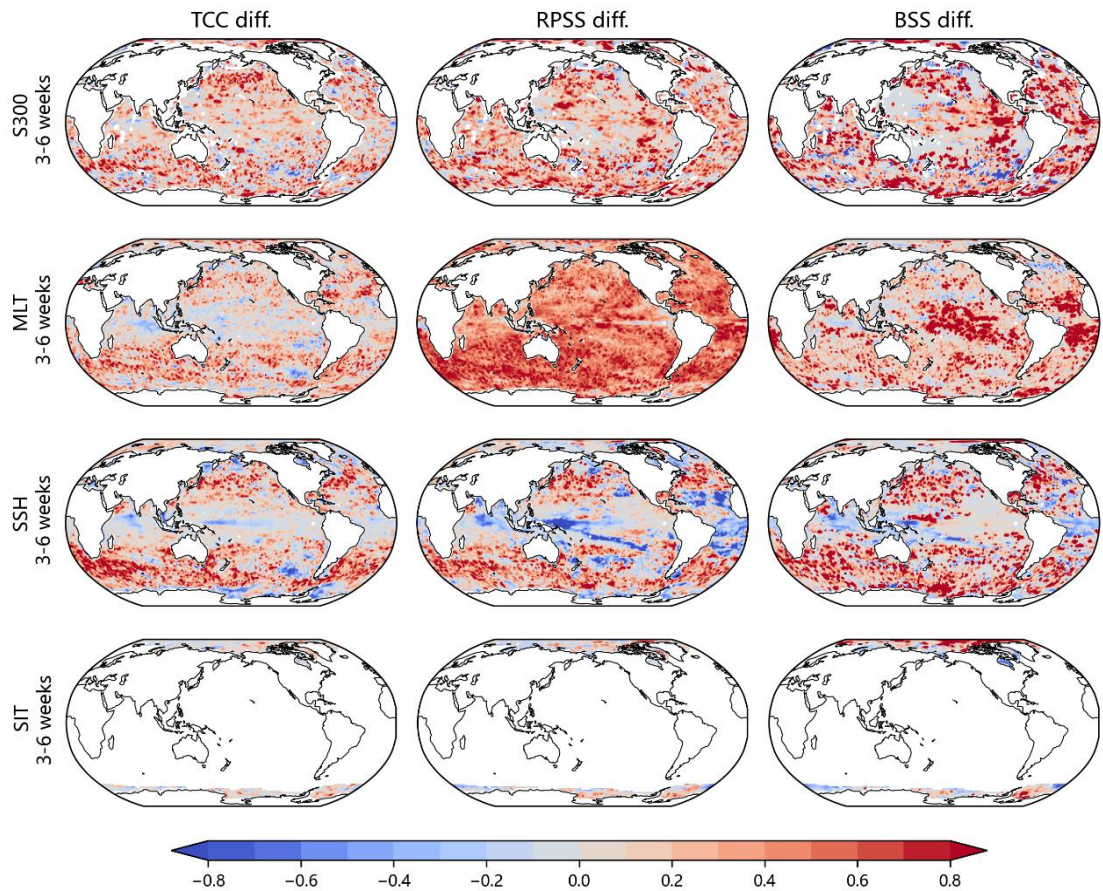

**Supplementary Figure 8:** Spatial maps of the differences in TCC (first column), RPSS (second column) and BSS (third column) between FengShun-CSM forecasts and ECMWF S2S forecasts for S300 (first row), MLT (second row), SSH (third row), and SIT (fourth row) at lead times of 3–6 weeks, using all testing data from 2021.

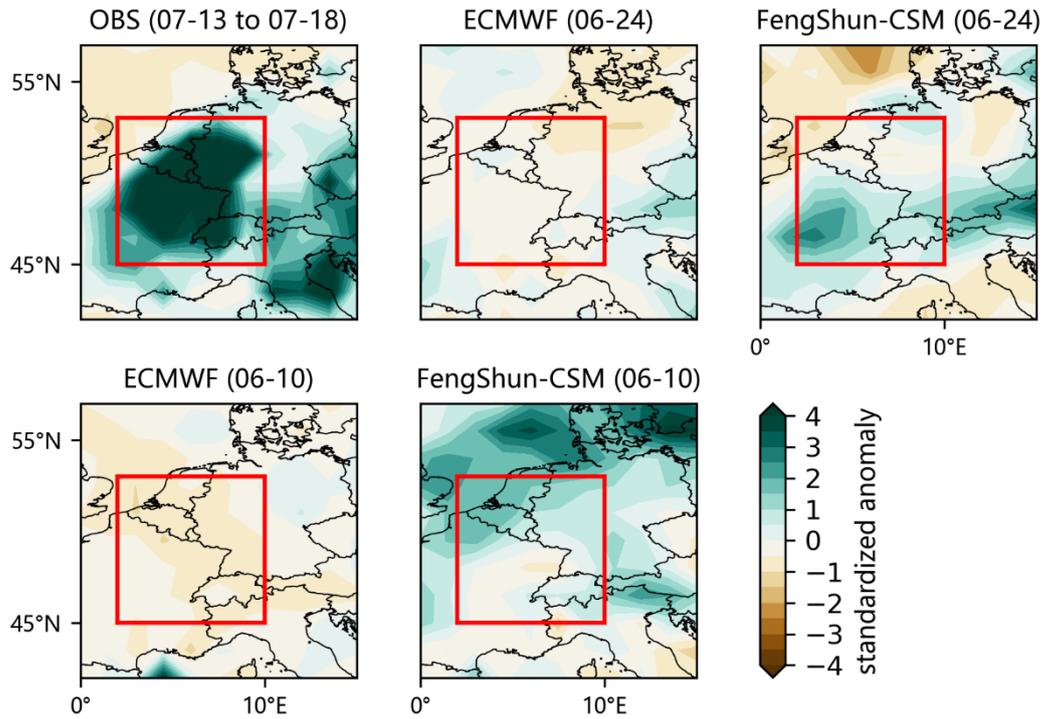

**Supplementary Figure 9:** Comparison of temporally averaged standardized TP anomaly maps for ERA5 observation from 13–18 July 2021 (first row, first column), ECMWF S2S initialized on 24 June (first row, second column), FengShun-CSM initialized on 24 June (first row, third column), ECMWF S2S initialized on 10 June (second row, first column), and FengShun-CSM initialized on 10 June (second row, second column). The red contour highlights the central region of the extreme precipitation event that occurred in central and western Europe in July 2021.

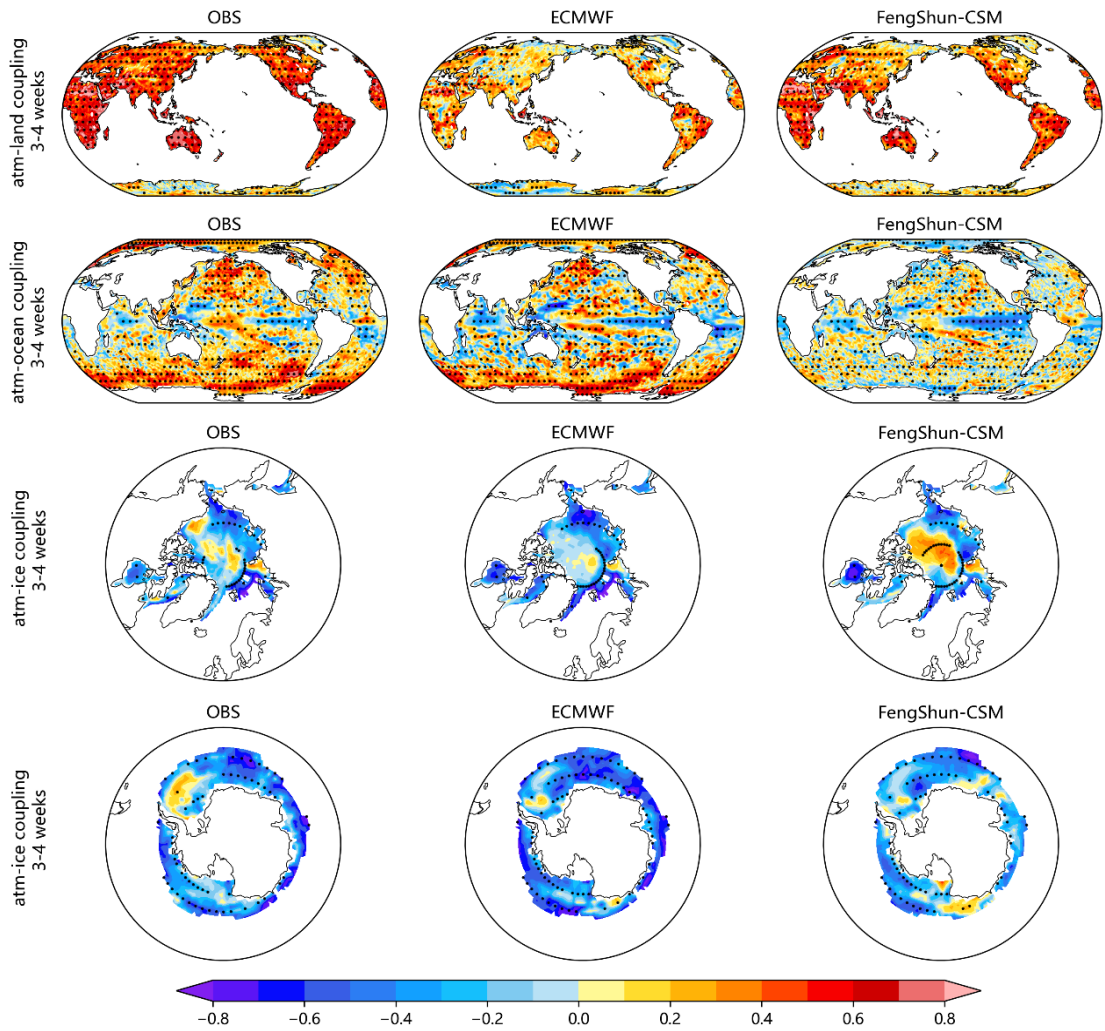

**Supplementary Figure 10:** Spatial maps of atmosphere-land coupling (first row), atmosphere-ocean coupling (second row), atmosphere-ice coupling over the Arctic region (third row), and atmosphere-ice coupling over the Antarctic region (fourth row) for OBS (first column), ECMWF S2S forecasts (second column), and FengShun-CSM forecasts (third column) at lead times of 3–4 weeks. The analysis is based on all testing data from 2021. In each subplot, the color scale represents the COR between different Earth system components. Black dotted areas indicate regions where the correlations are statistically significant at the 95% confidence level.

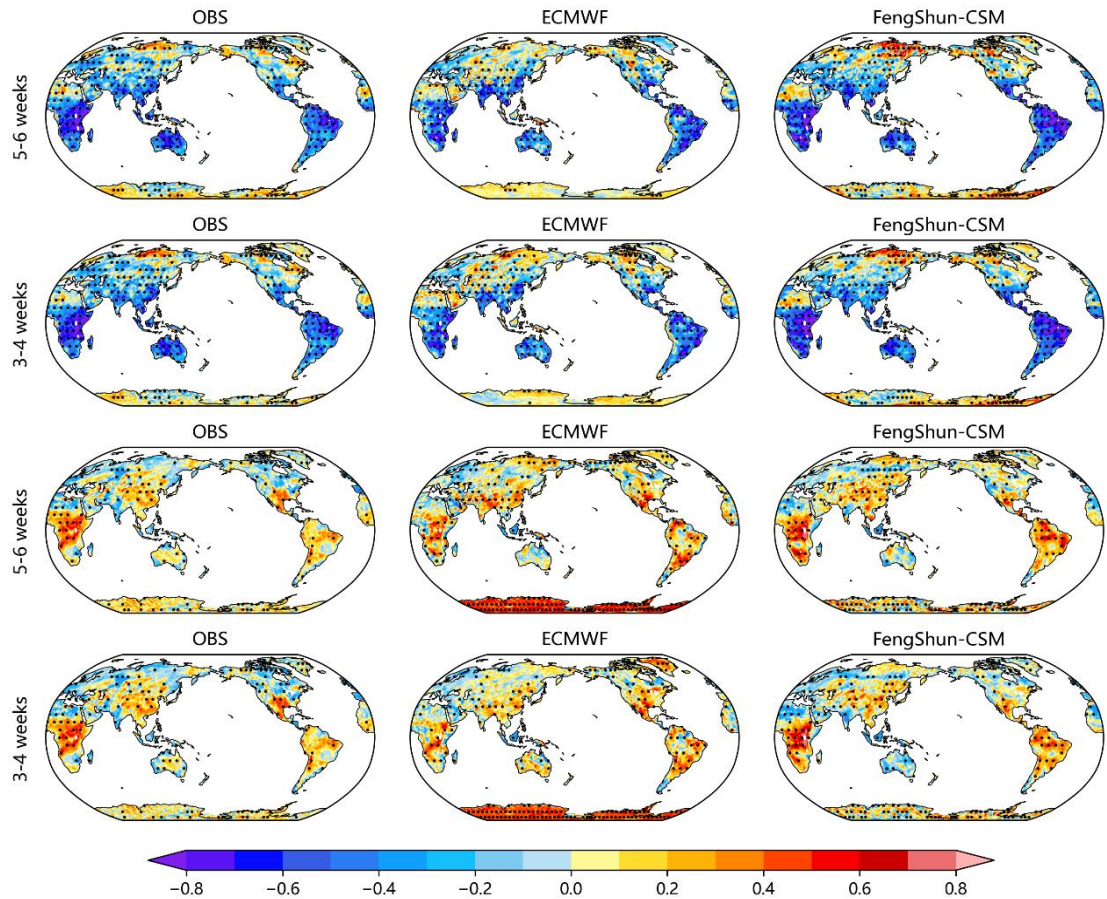

**Supplementary Figure 11:** Spatial maps of atmosphere-land coupling for OBS (first column), ECMWF S2S forecasts (second column), and FengShun-CSM forecasts (third column) at lead times of 5–6 weeks (first and third rows) and 3–4 weeks (second and fourth rows). The analysis is based on all testing data from 2021. In the subplots of the first and second rows, the color scale represents CORs between SM100 and T2M. In the subplots of the third and fourth rows, the color scale represents the CORs between SM100 and mean sea level pressure (MSL). Black dotted areas indicate regions where the correlations are statistically significant at the 95% confidence level.

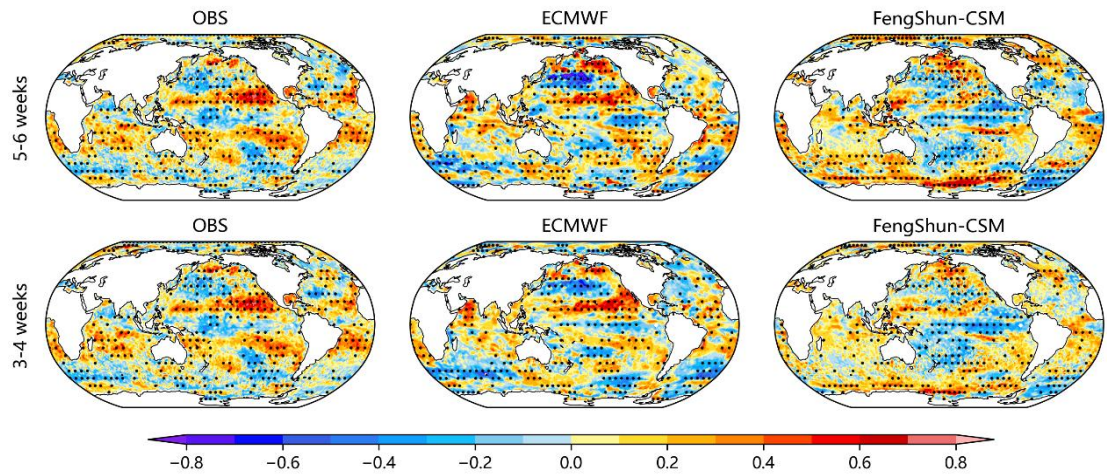

**Supplementary Figure 12:** Spatial maps of atmosphere-ocean coupling for OBS (first column), ECMWF S2S forecasts (second column), and FengShun-CSM forecasts (third column) at lead times of 5–6 weeks (first row) and 3–4 weeks (second row). The analysis is based on all testing data from 2021. In the subplots, the color scale represents CORs between MLT and MSL. Black dotted areas indicate regions where the correlations are statistically significant at the 95% confidence level.

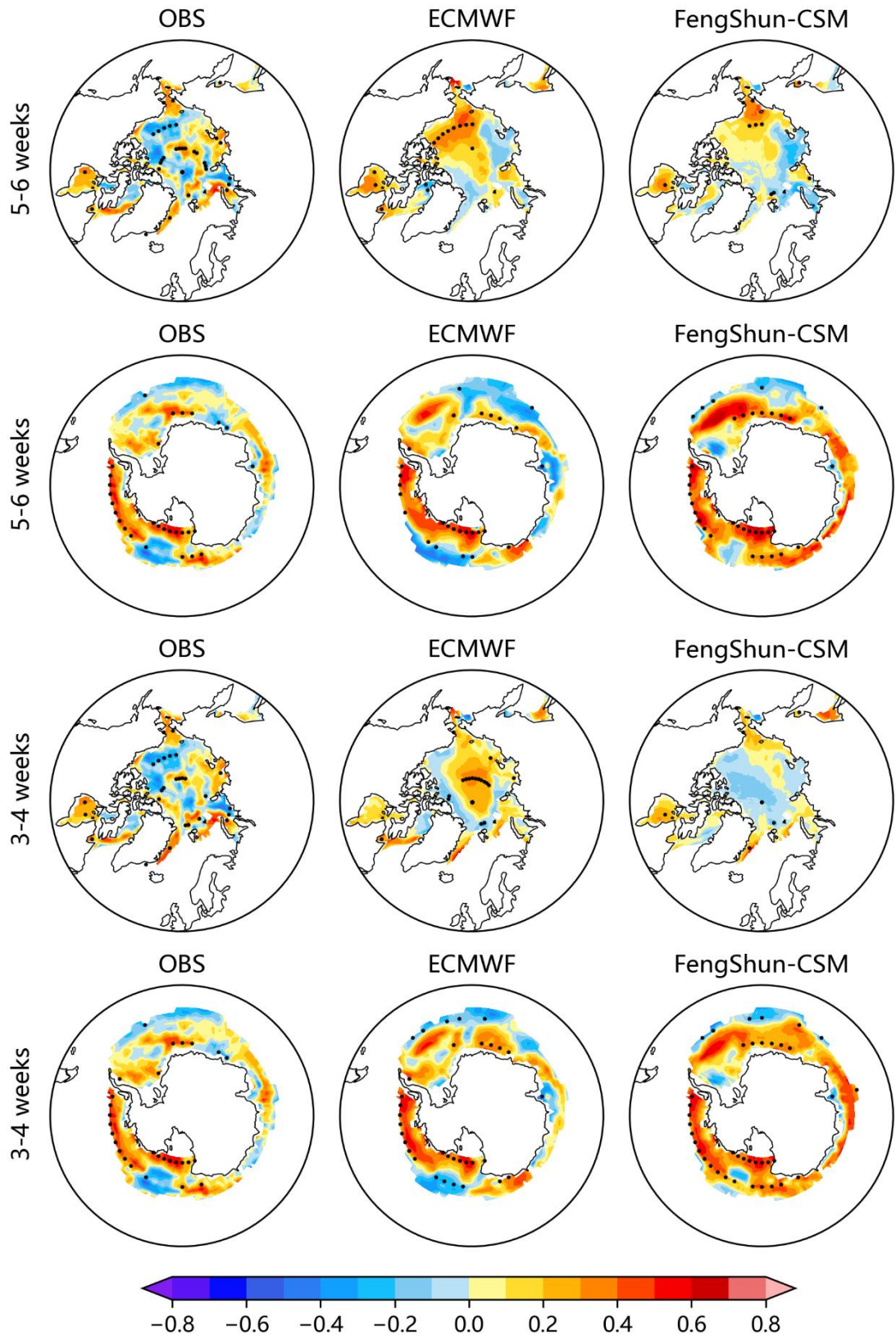

**Supplementary Figure 13:** Spatial maps of atmosphere-ice coupling over the Arctic and Antarctic regions for OBS (first column), ECMWF S2S forecasts (second column), and FengShun-CSM forecasts (third column) at lead times of 5–6 weeks (first and second rows) and 3–4 weeks (third and fourth rows). The analysis is based on all testing data from 2021. In the subplots, the color scale represents CORs between SIC and MSL. Black dotted areas indicate regions where the correlations are statistically significant at the 95% confidence level.

**Supplementary Table 1:** A summary of all variable names and their abbreviations.

| Type | Full name | Abbreviation |
|---|---|---|
| Upper-air variables | geopotential | Z |
| | temperature | T |
| | u component of wind | U |
| | v component of wind | V |
| | specific humidity | Q |
| Single-air variables | 2 m temperature | T2M |
| | minimum 2 m temperature | T2M_MIN |
| | maximum 2 m temperature | T2M_MAX |
| | 2 m dewpoint temperature | D2M |
| | 10 m u wind component | U10 |
| | 10 m v wind component | V10 |
| | 100 m u wind component | U100 |
| | 100 m v wind component | V100 |
| | mean sea-level pressure | MSL |
| | total cloud cover | TCC |
| | outgoing longwave radiation | OLR |
| | total precipitation | TP |
| Land variables | soil moisture from 0 to 289 cm | SM |
| | soil temperature from 0 to 289 cm | ST |
| | snow density | RSN |
| | snow depth | SD |
| Upper-ocean variables | potential temperature | PT |
| | salinity | S |
| | u component of surface current | OCU |
| | v component of surface current | OCV |
| Single-ocean variables | sea surface height | SSH |
| | ocean mixed layer thickness | MLT |
| Sea-ice variables | sea ice concentration | SIC |
| | sea ice thickness | SIT |